\documentclass{article}

\usepackage[preprint]{neurips_2025}

\usepackage[utf8]{inputenc} 
\usepackage[T1]{fontenc}    
\usepackage{hyperref}       
\usepackage{url}            
\usepackage{booktabs}       
\usepackage{amsfonts}       
\usepackage{nicefrac}       
\usepackage{microtype}      
\usepackage{xcolor}         

\usepackage{graphicx} 
\usepackage{amsmath} 
\usepackage{multirow} 
\usepackage{makecell} 
\usepackage{float}

\title{LLM world models are mental: Output layer evidence of brittle world model use in LLM mechanical reasoning}

 \author{
 	Cole Robertson \\
  	Department of Psychology \\
  	Emory University \\
  	Atlanta, GA 30322 \\
  	\texttt{cole.robertson@emory.edu}
  	\And 
  	Philip Wolff \\
  	Department of Psychology \\
  	Emory University \\
  	Atlanta, GA 30322 \\
  	\texttt{pwolff@emory.edu}
  }

\begin{document}

\maketitle

\begin{abstract}
  	Do large language models (LLMs) construct and manipulate internal ``world models,'' or do they rely solely on statistical associations represented as output layer token probabilities learned from data? We adapt cognitive science methodologies from human mental models research to test LLMs on pulley system problems using Ti\textit{k}Z-rendered stimuli. Study 1 examines whether LLMs can estimate mechanical advantage (MA) while distinguishing relevant from irrelevant system components and disregarding distractor elements. State-of-the-art models performed marginally but significantly above chance when exact estimate-label matches were required, and their estimates correlated significantly with ground-truth MA. Critically, tested models selectively attended to meaningful variables (e.g., number of ropes and pulleys) while ignoring features irrelevant to MA (e.g., rope diameter, pulley diameter, ceiling height). Significant correlations between number of pulleys and model estimates suggest that models employed a ``pulley counting'' heuristic to approximate MA, without necessarily simulating pulley systems to derive precise values. Study 2 tested this by probing whether LLMs represent global features crucial to MA estimation. Models evaluated a functionally connected pulley system against a ``fake'' system with randomly placed components. Without explicit cues, models identified the functional system as having greater MA with $F1=0.8$, suggesting LLMs could represent systems well enough to differentiate jumbled from functional systems. Study 3 built on this by asking LLMs to compare functional systems with matched systems which were ``connected up'' but which transferred no force to the weight; LLMs identified the functional system with $F1=0.46$, suggesting random guessing. Insofar as they may generalize\textemdash which is not tested\textemdash these findings are compatible with the notion that LLMs manipulate internal ``world models'' analogous to human mental models, sufficient to exploit statistical associations between pulley count and MA (Study 1), and to approximately represent system components' spatial relations (Study 2). However, they may lack the representational fidelity to represent and reason over nuanced structural connectivity (Study 3). We conclude by advocating for the utility of cognitive scientific methods developed to investigate human mental models in interrogating the world-modeling capacities of artificial intelligence systems.
\end{abstract}

\section{Introduction}

The rapid advancement of large language models (LLMs) has raised fundamental questions about the nature and extent of their intelligence. These systems show signs of constructing and manipulating internal models of the world, but how robust are they? Do they, to some extent, engage in flexible, generalizable reasoning? Or are their successes best understood as statistical pattern matching, relying on surface-level correlations learned from training distributions, without deeper comprehension? How can we tell the difference? Beginning to evidence answers to these questions carries significant implications for the trajectory of artificial intelligence (AI), particularly with respect to the feasibility of achieving artificial general intelligence (AGI).  
open
One perspective suggests that the recent generation of LLMs exhibits emergent cognitive capacities that go beyond mere next-token prediction. This claim is supported by findings that models such as GPT-4 can successfully navigate complex, multi-step reasoning tasks, solve novel problems in diverse domains, and demonstrate inferential processes over world models that extend beyond their training data \cite{Bubeck2023}. From this standpoint, the trajectory of LLM development may be seen as a path toward AGI, with large-scale pretraining yielding representations that approximate world models.  

An alternative perspective holds that LLMs, despite their impressive performance, do not engage in genuine reasoning but instead rely \textbf{only} on statistical regularities learned from vast amounts of data. Under this view, even seemingly sophisticated inferences can often be traced back to approximate retrieval from training distributions, rather than the kind of flexible, generalizable reasoning over robust ``mental models'' of the world that characterizes human cognition \cite{Marcus1998}. Recent empirical work further challenges the robustness of LLM reasoning, demonstrating that when tasks are carefully modified to fall outside typical training distributions, model performance deteriorates more sharply than does human performance \cite{Lewis2024}. For instance, simple tactics that do not even beat human amateurs were found to beat superhuman Go AIs \cite{Wang2023}, suggesting AIs lacked deep understanding of concepts fundamental to Go, despite superhuman performance. This suggests that while deep neural nets like LLMs may exhibit apparent ability to reason flexibly about novel problems, these abilities are often fragile and dependent on superficial features rather than deep understanding. This has led to calls in the literature to integrate configurable, predictive world models into AI systems to make them robust enough to use in critical tasks \cite{LeCun2022, Marcus2020}. 

Could such models, to some extent, be an emergent quality of the deep neural nets which already exist? A critical test for evaluating this question involves assessing performance on out-of-distribution (OOD) problems—tasks designed to require generalization beyond their training data. In cognitive science, similar tests have been used to investigate human reasoning, particularly in the study of mental models—internal representations that allow individuals to simulate and predict the behavior of physical systems \cite{Hegarty2004}. The ability to construct and manipulate such models is central to human cognition, allowing for flexible problem solving in novel contexts. If LLMs exhibit similar capacities, their outputs should reflect systematic reasoning over such ``mental'' representations of the world rather than statistical artifacts.  

\subsection{Mental models in human cognition and AI}

The study of mental models has played a foundational role in cognitive science, offering a framework for understanding how individuals reason about mechanical systems and the physical world. Classic work in this domain has demonstrated that humans often engage in mental simulation, constructing internal representations that allow them to predict outcomes and infer causal relationships \cite{ Hegarty1988, Schwartz1996}. Research on mechanical reasoning has shown that people mentally animate physical systems, decomposing their interactions into step-wise sequences that mirror real-world dynamics \cite{Hegarty2004}. Similarly, studies on spatial cognition indicate that mental rotation and analog imagery serve as core mechanisms for manipulating representations of physical objects and thereby making predictions about the world \cite{Shepard1971}. 

This perspective has begun to inform investigations of reasoning in LLMs. Recent work suggests that LLMs may implicitly encode and reason over world models which may be analogous to human mental simulation. For instance, studies have explored how LLMs reason about embodied experiences and infer causal relationships from textual inputs \cite{Xiang2023}. Other research has proposed that integrating world models\textemdash internal representations of social, physical, or geographical phenomena\textemdash could enhance LLMs’ ability to generalize across contexts \cite{Hu2023}. However, existing studies also highlight significant limitations: LLMs frequently struggle with analogical generalization \cite{Lewis2024}. Such findings evidence critiques of ML approaches to AI which acknowledge impressive progress but argue that deep neural systems like LLMs lack robustness \cite{Marcus2020}, and argue AI researchers should look to explicitly incorporate elements of human cognitive models to develop more flexible, trustworthy, and robust world modeling capabilities in AI systems \cite{Marcus2020, Jabarian2024}.

If AI systems are to approximate human-like general intelligence, the development of robust internal world models\textemdash whether explicitly encoded, learned from data, or emergent from network scale\textemdash may be essential. This requires developing methods to characterize latent internal world models which may be represented in LLM network weights, and disentangle such world models from output layer token probabilities. Methods to evaluate whether deep networks reason over internally simulated world models, and determine the conditions under which their representations may break down, may help achieve progress toward robust AI.

\begin{figure}[t]
	\centering
	\includegraphics[width=0.9\columnwidth, keepaspectratio]{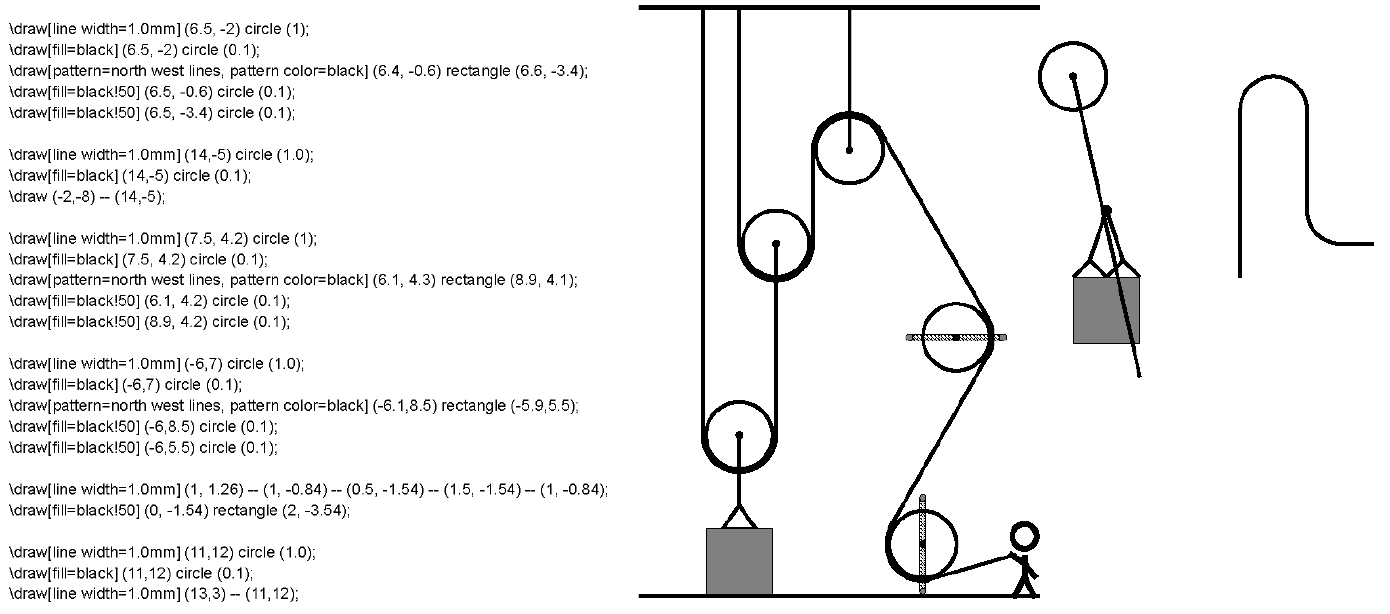}
	\caption{Example of a system with $MA=4$, as Ti\textit{k}Z source code excerpt (\textit{left}) and rendered as a PDF (\textit{right}). In all studies, LLMs were only shown Ti\textit{k}Z source code as in \textit{left}.} 
	\label{fig:s1_diagram}
\end{figure}

\subsection{Evaluating LLM world models with pulley systems}  

One approach to evaluating human reasoning over mental models involves assessing ability to infer the mechanical advantage of pulley systems  \cite{Hegarty1994, Hegarty1997, Hegarty2004}. Pulley systems are configurations of ropes and pulleys that reduce the force needed to lift heavy objects. Mechanical advantage (MA) quantifies this force reduction and is defined as the ratio between the distance over which input force is applied and the resulting displacement of the load. In an idealized system, MA determines the proportional decrease in force required to lift a weight, meaning that with an MA of 2, the applied force is halved, but the rope must be pulled twice the distance the weight moves. Pulley systems provide a reasonable test-bed to evaluate LLM reasoning over world models for several reasons. First, they have been used to investigate how people construct mental representations of mechanical systems, making them a well-established benchmark for studying mental simulation of physical phenomena. Second, pulley systems exhibit discrete complexity\textemdash there is no one-to-one mapping between individual elements and MA, requiring an integrated understanding of how MA emerges from particular arrangements of system components. Third, solving pulley problems correctly requires the ability to identify relevant system components while disregarding irrelevant features, offering a way to assess selective attention in LLM inference.  

Moreover, this task presents an opportunity to take LLMs outside their training distribution. While LLMs have encountered vast amounts of text, we deem it unlikely that they have been systematically trained on pulley system reasoning in the precise form presented in this study—particularly reasoning over pulley systems represented as Ti\textit{k}Z  \cite{Tantau2024} source code, a \TeX -based markup language for creating diagrams (see Figure~\ref{fig:s1_diagram}). This aligns with emphasis in the literature on using OOD problems to test whether reasoning abilities are generalized, allowing researchers to determine whether AI models generalize underlying principles rather than relying on surface-level retrieval. Given previous findings that LLMs often fail when reasoning tasks are slightly modified \cite{Lewis2024}, this study offers a test of whether LLMs, to some degree, construct world models and make inferences from reasoning systematically over them.  

This study investigates LLM performance on pulley systems to contribute to ongoing discussions about robust AI, the extent of emergent reasoning over world models in LLMs, and the broader question of LLMs' progression toward AGI.

\texttt{}\section{Study 1}

\begin{figure}[t]
	\centering
	\includegraphics[width=\columnwidth, keepaspectratio]{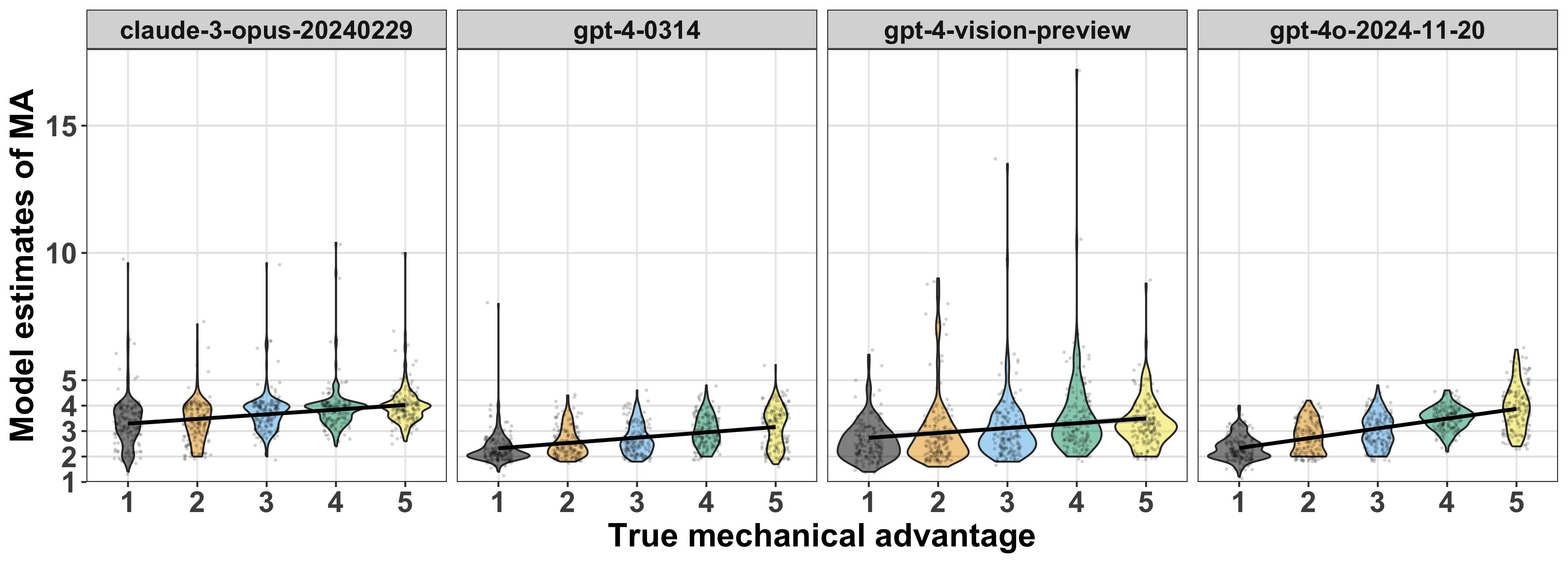}
	\caption{Mean model estimates of MA over true MA.} 
	\label{fig:s1_violin}
\end{figure}

Study 1 had two aims. First, it was designed to establish whether state-of-the-art LLMs could infer mechanical advantage from diagrams of pulley systems represented in Ti\textit{k}Z source code. Second, it was designed to establish whether LLMs could attend to system features which are correlated with mechanical advantage (e.g. number of ropes, number of pulleys) while disregarding irrelevant features (e.g. rope diameter, pulley radius, ceiling height), as well as distractor elements. This dual approach is based directly on empirical work with human subjects which establishes the ability to attend selectively to relevant diagram features as a critical determinant of performance \cite{Hegarty1988}. If LLMs are able to represent internal models of mechanical systems, then they should be able to identify which elements of a scene are relevant and which elements are irrelevant. Additionally, they should be able to generate MA estimates that correlate with ground truth.

\subsection{Methods and materials}

\paragraph{Data collection:}

Four LLMs completed the experiment: \textit{claude-3-opus-20240229} \cite{Anthropic2024}, \textit{gpt-4-0314} \cite{OpenAI2023}, \textit{gpt-4o-2024-11-20} \cite{OpenAI2024}, and \textit{gpt-4-vision-preview} \cite{OpenAI2023a}. These  were selected because at the time of the experiment they represented the state-of-the-art. Additionally, in the case of the first three, weight-frozen checkpoints were available, which is important for reproducibility.  Inference was performed through paid-access APIs available from OpenAI and Anthropic, February\textendash March 2024 (May 2025 for \textit{gpt-4o-2024-11-20}). For all models, inference hyperparameters were $temperature=0.5$, $max\ tokens=2500$, and for the GPT models, $top\ p=1$, $frequency\ penalty=0.0$, and $presence\ penalty=0.0$ (the last three were not available from the Anthropic APIs at the time of the experiment).

\paragraph{Materials:}

Materials consisted of pulley system diagrams created in the Ti\textit{k}Z markup language. There were five factors. First, \textit{mechanical advantage} (MA). Two different systems were created for each of $n=5$ MA levels (1, 2, 3, 4, 5). Second, \textit{number of pulleys}. Since the number of pulleys strongly correlates with MA, we introduced $n=3$ variants per system by adding irrelevant redirect pulleys (see Figure~\ref{fig:s1_diagram}), preventing a simple 1:1 mapping. This yielded $n=30$ system variants. Three additional factors, unrelated to MA, were fully crossed over these $n=30$ systems: \textit{ceiling height} ($n=3$ levels), \textit{pulley radius} ($n=3$), and \textit{rope diameter} ($n=3$), resulting in $n=810$ unique diagrams. Since stimuli were presented as Ti\textit{k}Z code, it was important there be no trivial 1:1 mapping between line count and system MA. Distractor elements were added\textemdash code blocks mimicking system components but not connected to the actual system. Placement and selection of these distractors were randomized, and lines were added until $n_{\text{lines}} = \max(n_{\text{lines}})$. An additional random number drawn from $\text{Uniform}\{0,10\}$ was then appended to each diagram to introduce stochasticity (see Figure~\ref{fig:s1_diagram}). Materials are available upon request by interested researchers. They are not public, as this would risk them being incorporated into language model training data scraped from the internet, which would obviate their ability to take models out of distribution.

\paragraph{Procedure:} 

Each diagram was presented $n=5$ times, for a total of $n=4050$ trials per LLM. Prompts were as follows: ``I am going to give you some Ti\textit{k}Z code that specifies a pulley system. I want you to estimate the mechanical advantage of the pulley system depicted in Ti\textit{k}Z code. I'm not asking you to compile the code, rather just analyze it and base your conclusions from the code directly. Each system element (rope, pulley, weight, supporting structures [ceiling, floor], and the person) are depicted in their own code block, separated by a blank line. In general, the code moves from top to bottom and from left to right. Some lines of code define distractor elements that are unrelated to the functional components of the pulley system. Your job is to differentiate these from the system, and then tell me the mechanical advantage of the system. Here is the Ti\textit{k}Z code depicting a pulley system \`{}\`{}\`{}Latex \{Ti\textit{k}Z code here\}\`{}\`{}\`{}. Please estimate the mechanical advantage of the system. In providing your answer, you can assume that the system is idealized, with no rope stretch and no friction. Feel free to outline your thinking in a step by step way. You MUST estimate the MA of the system, even if it is unclear to you.  What is the mechanical advantage of the system depicted in the Ti\textit{k}Z code above?'' Order of the trials was randomized, and no feedback was given. Experimental scripts are available at \url{https://osf.io/mn9hy/}.

\paragraph{Data processing:} 

A data cleaning step was necessary. \textit{GPT-4} was instructed to extract numerical MA from each response (``estimated MA'' ), which was compared with $N=101$ human responses from a research assistant on the same data-cleaning task. Cohen's kappa indicated near-perfect agreement, $\kappa = 0.99$, $z=29.8$, $p < .001$ \cite{Landis1977}. This process yielded $N=4050$ observations (\textit{gpt-4-0314} and \textit{claude-3-opus-20240229}), $N=3848$ (\textit{gpt-4-vision-preview}), and $n=4049$ (\textit{gpt-4o-2024-11-20}).

\begin{figure}[t]
	\centering
	\includegraphics[width=\columnwidth, keepaspectratio]{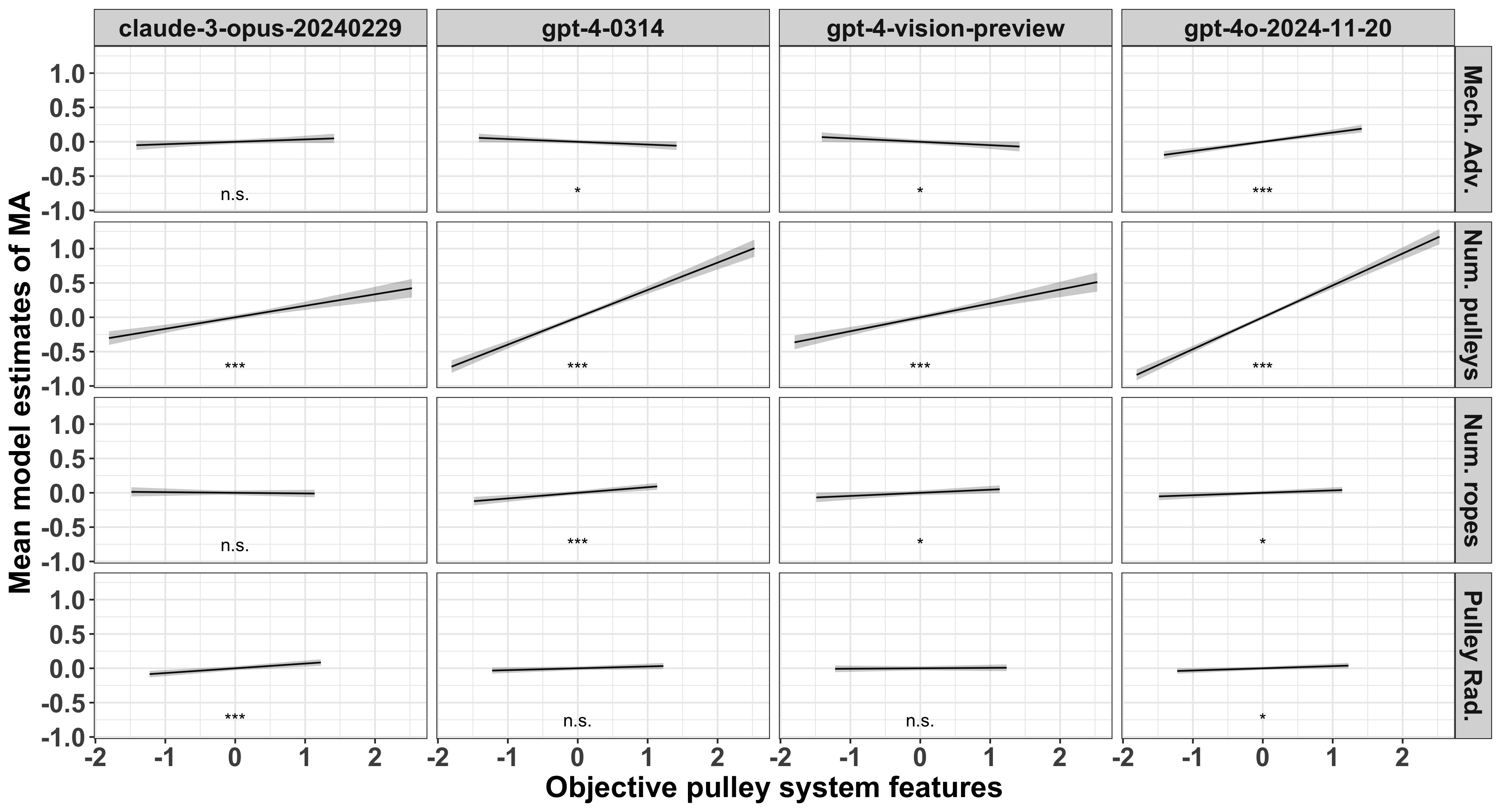}
	\caption{Model estimates of MA regressed over objective pulley system features. Features where all $\beta$s were n.s. are not shown (\textit{number of code lines, rope width, ceiling height}, see Technical Appendix, Section~\ref{sec:tech_extra}). Signif. codes: *** p\textless0.001, ** p\textless0.01, * p\textless0.05.
	} 
	\label{fig:s1_predict}
\end{figure}

\subsection{Results and discussion}

Compared to an expected proportion of 20\% (which is conservative, as models were not informed that true mechanical advantage values ranged from 1\textendash 5), all models identified the correct MA at rates significantly above chance: \textit{gpt-4o-2024-11-20}: 26.1\% [95\% CI: 25.0, 100], \textit{claude-3-opus-20240229}: 23.7\% [95\% CI: 22.6, 100], \textit{gpt-4-0314}: 23.4\% [95\% CI: 22.3, 100], and \textit{gpt-4-vision-preview}: 23.1\% [95\% CI: 22.0, 100]; all $p < .001$. However, though significantly better than chance, most answers were incorrect. Did they nonetheless tend to correlate with ground truth? To test this we calculated the mean over the $n=5$ trials for each diagram, $\bar{X}_{m,d} = \frac{1}{n} \sum_{i=1}^{n} X_{m,d,i}$, the mean for model $ \mathbf{m} $ on diagram $ \mathbf{d}$, across $\mathbf{i}$ trials under the same conditions. We then regressed the result over ground truth MA using ordinary least squares, $\bar{X}_{m,d} = \beta_0 + \beta_1 \text{MA} + \varepsilon$. For \textit{gpt-4-0314}, MA significantly predicted responses, $ \beta = 0.424$, $SE = 0.032$, $t(808) = 13.3$, $p < .001$, $R^2 = .18$. For \textit{gpt-4-vision-preview}, MA was also significant, though weaker: $\beta = 0.20$, $SE = 0.035$, $t(808) = 5.80$, $p < .001$, $R^2 = .04$. For \textit{claude-3-opus-20240229}, the relationship was intermediate in strength: $\beta = 0.283$, $SE = 0.034$, $t(808) = 8.39$, $p < .001$, $R^2 = .08$. Finally, for \textit{gpt-4o-2024-11-20}, the relationship was strongest: $\beta = 0.651$, $SE = 0.027$, $t(808) = 24.39$, $p < .001$, $R^2 = .42 $. These results indicate that model estimates tend to correlate with true MA, even when not precisely correct, and especially for \textit{gpt-4o-2024-11-20}, for which we found the strongest associations (see Figure~\ref{fig:s1_violin}), and Technical Appendix Table~\ref{tab:s2-ma-regression-results}.

Finally, we wanted to establish whether models were able to selectively attend to relevant features of the pulley diagrams.  We estimated the following OLS regression: $\bar{X}_{m,d} = \beta_0 + \mathbf{Z} \boldsymbol{\beta} + \varepsilon$, where $ \mathbf{Z} $ is the vector of predictor variables (see Figure~\ref{fig:s1_predict}). Number of pulleys was the strongest predictor of model estimates. For the GPT models, the number of supporting ropes also predicted mean estimated MA, and pulley radius was sometimes significant but appeared to be weak, e.g. for \textit{Claude}, the standardized coefficient ($\beta = 0.069$) was markedly smaller than that of number of pulleys ($\beta = 0.168$), indicating that pulley radius contributed relatively little to model predictions compared to number of pulleys, and should be interpreted as a weak effect in practical terms, see Technical Appendix Table~\ref{tab:s2-regression-results}. Crucially, the coefficient for MA was either negative or non-significant for all models except \textit{gpt-4o-2024-11-20} when controlling for other system features. The strong relationship between pulley count and estimated MA indicates that these models employed a ``pulley counting'' strategy to approximate MA. Notably, this count refers to \textit{system} pulleys, excluding random distractors, suggesting models could differentiate between functional and irrelevant pulleys. Interestingly, the GPT models appeared to use a more nuanced strategy, incorporating both pulley and supporting rope counts. This approach is subtly complex since supporting rope count can be a better predictor of MA than pulley count, especially in cases with non-contributory redirects (see Figure~\ref{fig:s1_diagram}). Of note is the positive $\beta$ coefficient between MA and model estimates for \textit{gpt-4o-2024-11-20} when controlling for other diagram features. This suggests \textit{gpt-4o-2024-11-20} was, to some degree, attending to pulley system arrangements in addition to employing proximate heuristics based on pulley count, which may explain \textit{gpt-4o-2024-11-20}'s better performance (26.1\% accuracy compared to ~23.7\% for the next best model).

Study 1 demonstrated that models could approximate MA from complex Ti\textit{k}Z-rendered pulley diagrams, differentiate between relevant and irrelevant features, and distinguish distractor elements from system-critical code. These findings are consistent with the view that models were flexibly reasoning over internal world models, but fell short of demonstrating this conclusively. Rather, the results suggest LLMs relied on simple pulley-counting heuristics\textemdash though heuristics which are based on an accurate model of the world, since the number of pulleys genuinely correlates with MA. Were the models truly attending to the structural arrangement of components that determines mechanical advantage? Study 2 directly investigated this question.

\section{Study 2}

Study 2 aimed to investigate if models were able to represent the connections between components of a pulley system\textemdash a crucial component of mechanical advantage.

\subsection{Methods and materials}

\paragraph{Data gathering:}

LLMs and inference procedures were the same as Study 1. Data were collected March\textendash April 2024 (May 2025 for \textit{gpt-4o-2024-11-20}).

\paragraph{Materials:}

\begin{figure}[t]
	\centering
	\includegraphics[width=0.9\columnwidth, keepaspectratio]{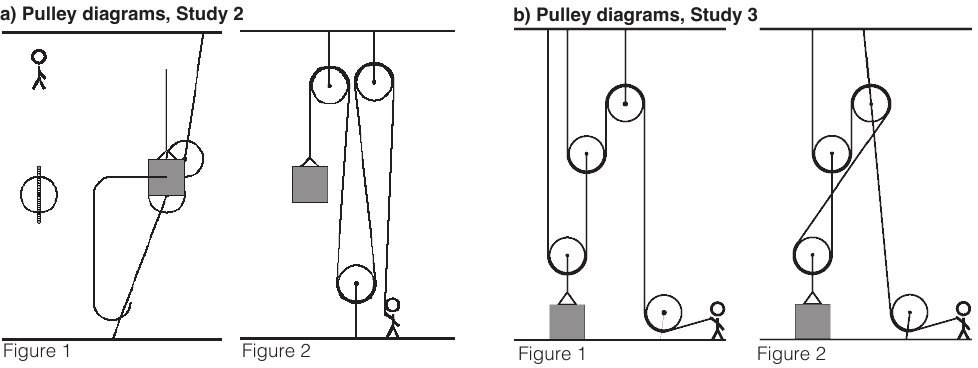}
	\caption{a) Study 2 stimuli: non-functional (\textit{left}) and functional (\textit{right}) paired pulley systems; b) In Study 3, non-functional diagrams were ``connected up'' but transferred no force to the weight.} 
	\label{fig:s2_diagram}
\end{figure}

As in Study 1, materials were pulley system diagrams created in Ti\textit{k}Z. However, the paradigm was altered. In Study 2, each of the $N=30$ pulley diagram variants  was paired with a non-functional diagram that was precisely matched for the number of pulleys, weights, human figures, and lines of Ti\textit{k}Z code, but in which system components were randomly placed. This resulted in $N=30$ unique paired diagrams, each comprising one real and one non-functional system (see Figure~\ref{fig:s2_diagram}). The placement of the functional system was randomized (i.e. left vs. right).

\paragraph{Procedure:}

Each diagram was presented $n=5$ times, for a total of $n=150$ trials per LLM. Prompts were as follows: ``I am going to give you some Ti\textit{k}Z code that specifies two pulley systems. I want you to tell me which one has a greater mechanical advantage. Each system element (rope, pulley, weight, supporting structures [ceiling, floor], and the person) is depicted in its own code block, separated by a blank line. In general, the code moves from top to bottom and from left to right. There are two systems -- Figure 1 (left) and. Figure 2 (right). Here is the Ti\textit{k}Z code depicting two pulley systems: \`{}\`{}\`{}Latex \{Ti\textit{k}Z code here\}\`{}\`{}\`{}. In providing your answer, you can assume that the systems are both idealized, with no rope stretch and no friction. Feel free to outline your thinking in a step by step way. Which system (Figure 1 or Figure 2) has the greater mechanical advantage? You MUST estimate which system has greater MA, even if it is unclear. Please do not equivocate or ask for more information.'' Order of the trials was randomized, and no feedback was given. 

\paragraph{Data processing:}

As in Study 1, a data cleaning step was needed. We instructed  \textit{gpt-4} to return ``left'' if a model had indicated Figure 1, and ``right'' for Figure 2. Given strong Study 1 results, we deemed further reliability analyses unnecessary. The resultant values were used to calculate whether model responses were correct (1) or incorrect (0). Due to non-compliance with prompts, this procedure resulted in $n = 150$ observations (\textit{claude-3-opus-20240229}); $n=149$ (\textit{gpt-4-vision-preview}); $n=143$ \textit{gpt-4o-2024-11-20}; and $n=146$ (\textit{gpt-4-0314}).

\subsection{Results and discussion}

Models demonstrated a mean accuracy of 80.7\% in correctly identifying the functionally connected system's greater MA (Technical Appendix, Table~\ref{tab:combined_metrics}). All models significantly surpassed the expected 50/50 baseline, $p < .001$ (Technical Appendix, Table~\ref{tab:combined-binom-results}). Perhaps surprisingly \textit{gpt-4o-2024-11-20} performed the worst on this task, despite performing the best in Study 1. Why this was the case is perplexing; if it struggled with the omitted information in the prompt (i.e. not being informed one system was non-functional), this suggests its representations of pulley systems are brittle enough to be seriously impacted by surface-level task details, affirming critiques of deep ML approaches to AI which foreground such issues \cite{Marcus2020}. Notwithstanding these complexities, the results suggest models were able to represent gestalt features of how each system was ``connected up'',  at least enough to differentiate between functional systems and jumbled non-connected ones. 

A qualitative analysis revealed that when models supplied explanations, they tended to justify their inferences in terms of hallucinated system features which indicated both systems were functional. An example response in which \textit{gpt-4-0314} characterizes a pulley system in Figure~\ref{fig:s2_diagram}a typifies the larger sample: ``Based on the provided Ti\textit{k}Z code, Figure 2 has a greater mechanical advantage. This is because it has more pulleys and a longer path for the rope, which would result in a greater mechanical advantage in an idealized system.'' Note, the number of pulleys in each of the panels in Figure~\ref{fig:s2_diagram}a is the same (3), and the premise that rope length has any bearing on MA is incorrect under idealized conditions. GPT is nonetheless correct: The left panel in Figure~\ref{fig:s2_diagram} confers no MA at all. This suggests that the LLMs sampled in Study 2 reasoned over latent representations of pulley systems, but that these may have been conceptually uncoupled from explicit reasoning at the output layer. 

Study 2 strengthens the findings from Study 1 suggesting that tested LLMs have some ability to reason over pulley systems, at least enough to represent the obvious disconnections between system components in the jumbled non-functional stimuli.

\section{Study 3}

Were LLMs genuinely representing and reasoning over pulley systems, or were they only picking up on simple cues of disorganization in the jumbled stimuli set in Study 2? Study 3 tested this using a subtle alteration of the stimuli\textemdash rather than creating non-functional systems by randomly placing system components, non-functional systems were created that were still ``connected up'' in a seemingly functional manner, but in such a way as to transfer no force to the weight (see Figure~\ref{fig:s2_diagram}b). If models were able to represent and reason about the connections between components of a pulley system\textemdash a crucial component of estimating mechanical advantage\textemdash they should be able to differentiate between these and genuinely functional systems.

\subsection{Methods and materials}

\paragraph{Data gathering:}

The LLMs tested were \textit{gpt-4o-2024-11-20} and \textit{claude-3-opus-20240229}, because frozen checkpoints for the other GPT models were no longer available at the time of Study 3. Inference hyperparameters were the same as Studies 1 \& 2. Data were collected in May 2025. 

\paragraph{Materials:}

As in Study 2, each of the $N=30$ pulley diagrams was paired with a non-functional system. This time, the non-functional diagrams were based directly on their functional counterparts, but altered such that pulling the rope transferred no force to the weight (see Figure~\ref{fig:s2_diagram}b). Non-functional systems were precisely matched for the number of pulleys, weights, human figures, and lines of Ti\textit{k}Z code\textemdash only the arrangement of components was changed. This resulted in $N=30$ paired diagrams, each comprising one real and one connected but non-functional system. 

\paragraph{Procedure:}

Each diagram pair was presented $n=10$ times ($n=5$ with the functional system on the left, and $n=5$ reversed), for a total of $n=300$ trials per LLM. Prompts were as follows: ``I am going to give you some Ti\textit{k}Z code that specifies two pulley systems. I want you to tell me which one has a greater mechanical advantage. Each system element (rope, pulley, weight, supporting structures [ceiling, floor], and the person) is depicted in its own code block, separated by a blank line. In general, the code moves from top to bottom and from left to right. There are two systems -- Figure 1 (left) and Figure 2 (right). Here is the Ti\textit{k}Z code depicting two pulley systems: \`{}\`{}\`{}Latex \{Ti\textit{k}Z code here\}\`{}\`{}\`{}. In providing your answer, you can assume that the systems are both idealized, with no rope stretch and no friction. Feel free to outline your thinking in a step by step way.  When the figure pulls on the rope, which system will move the block with LESS force? You MUST provide an answer, either left or right, even if it is unclear. Please do not equivocate or ask for more information.'' Order of the trials was randomized, and no feedback was given. 

\paragraph{Data processing:}

Data cleaning was identical to Study 2, resulting in $n=300$ for both models. Trials where LLMs asserted that MA was equal between the figures were counted as failures.

\subsection{Results and discussion}

Models demonstrated a mean accuracy of 50.8\% in correctly identifying the functionally connected system's greater MA (Technical Appendix, Table~\ref{tab:combined_metrics}), no better than chance. The model \textit{gpt-4o-2024-11-20} achieved an accuracy of $0.553$, better than chance using a Binomial exact test,  $95\% CI = [0.504, 1]$, $n = 300$, $p= .0367$, though only marginally. The model \textit{claude-3-opus-20240229} did not perform above chance levels, $p = .908$ (see Technical Appendix Table~\ref{tab:combined-binom-results}). Did GPT's statistically significant but poor performance indicate it understood some of the pulley systems well enough to differentiate between functional and non-functional ones? If so, its answers to mirrored diagrams should correlate positively, i.e. it should score equally as well (or poorly) on a paired contrast when the functional system is left or right. In fact, the correlation was negative and non-significant, \textit{gpt-4o-2024-11-20}, $r(28) = -0.28$, $p = .138$, $95\% CI =  [-0.58, 0.09]$; \textit{claude-3-opus-20240229} was no better, $r(28) = -0.28$, $p = .128$, $95\% CI  = [-0.58, 0.08]$. This suggests GPTs above-chance performance was a statistical artifact, rather than an indication it could represent pulley system arrangements with sufficient fidelity to perform above chance levels.

Study 3 indicates a clear limit to tested models' ability to reason over latent representations of pulley systems, at least as presented in these Ti\textit{k}Z-based stimuli. Tested models failed to understand and make correct inferences about a crucial determinant of mechanical advantage: How system components are arranged so as to transfer force to moving weight.

\section{General Discussion}
Our results were as predicted by the view that LLMs represent and reason over internal representations of pulley systems\textemdash to a point.  In Study 1, tested models selectively attended to relevant mechanical features while disregarding irrelevant ones, estimated MA with above-chance accuracy, and these estimates correlated with ground truth values. However, for all but one model (\textit{gpt-4o-2024-11-20}), the effect of true MA on model estimates was either negative or non-significant when controlling for other system features, notably number of pulleys. This suggests models relied on a pulley-counting heuristic to arrive at approximations of MA, an interpretation supported by notably low model accuracy (the best model \textit{gpt-4o-2024-11-20} was correct 26.1\% of the time). In Study 2, models distinguished functional from non-functional pulley systems at high accuracy, despite failing to articulate the basis for their judgments. LLMs may have represented system features (jumbled arrangements), but they failed to explicitly reason over them\textemdash suggesting that LLMs encode latent knowledge that influences their outputs but is not present in output layer tokens. However, Study 3 exposed critical limitations. When challenged with stimuli in which both diagrams were superficially well-formed yet only one was genuinely functional, model performance collapsed to chance. This suggests that previously observed successes were less driven by robust reasoning over high-fidelity latent models of pulley systems, and more by coarse heuristics or shallow cues. The lack of within-item consistency across mirrored trials further supports this interpretation, revealing that even statistically significant performance was unreliable and likely not underpinned by stable internal models.

\paragraph{Limitations:}
A central limitation is the uncertainty regarding whether tested pulley stimuli are truly out-of-distribution; exposure during training cannot be ruled out. Future work should further deconfound pulley count from MA and employ novel or synthetic diagramming languages to limit reliance on familiar formats like Ti\textit{k}Z. Extending the paradigm to other physical systems (e.g., levers, gear trains) may help assess the breadth of these internal representations and clarify their correspondence to mechanistic reasoning or more general world-model simulation abilities. The present ``cognitive'' approach is also limited\textemdash deconfounding output layer token representations from potential latent world models represented in network weights is difficult. Without combining the present methods with interpretability research which unpacks knowledge represented in deep network weights, it is difficult to be certain on the basis of output tokens whether and to what extent LLMs are reasoning over models of the world, or how robust those models may be. 

\paragraph{Conclusions:}
Our findings offer mixed support to the view that LLMs reason over world models in a way which is\textemdash to some degree\textemdash independent from output token probabilities. This aligns with claims that large-scale pretraining can lead to approximate world models \cite{Bubeck2023}. However, the brittleness of these representations, especially in Study 3, aligns with critiques emphasizing the lack of robustness and generality in current systems \cite{Marcus1998, Lewis2024}. While encouraging, our results underscore the need for refined stimuli and interpretability tools to trace how and when internal representations arise or fail. Integrating the present approach with interpretability research\textemdash such as efforts to extract conceptual features from LLMs' network circuitry \cite{Templeton2024}\textemdash may help elucidate the network circuits underlying any reasoning over world models in which AI systems may engage. This fusion promises a shift from the present cognitive approach to a neuro-cognitive science of AI, enabling targeted refinements of world model circuits and advancing the development of more robust AI systems.
\newpage

\bibliography{./LLM_models_are_mental-Robertson_et_al}

\begin{thebibliography}{22}
\providecommand{\natexlab}[1]{#1}
\providecommand{\url}[1]{\texttt{#1}}
\expandafter\ifx\csname urlstyle\endcsname\relax
  \providecommand{\doi}[1]{doi: #1}\else
  \providecommand{\doi}{doi: \begingroup \urlstyle{rm}\Url}\fi

\bibitem[{Adly Templeton et al.}(2024)]{Templeton2024}
{Adly Templeton et al.}
\newblock {Scaling monosemanticity: Extracting interpretable features from
  Claude 3 Sonnet}.
\newblock \emph{Transformer Circuits Thread}, 2024.

\bibitem[Anthropic(2024)]{Anthropic2024}
Anthropic.
\newblock {The Claude 3 Model Family: Opus, Sonnet, Haiku Anthropic}.
\newblock \emph{Anthropic}, 2024.

\bibitem[{Gary Marcus et al.}(1998)]{Marcus1998}
{Gary Marcus et al.}
\newblock {Rethinking eliminative connectionism}.
\newblock \emph{Cognitive Psychology}, 282\penalty0 (37):\penalty0 243--282,
  1998.
\newblock \doi{10.1006/cogp.1998.0694}.

\bibitem[Hegarty(2004)]{Hegarty2004}
Mary Hegarty.
\newblock {Mechanical reasoning by mental simulation}.
\newblock \emph{Trends in Cognitive Sciences}, 8\penalty0 (6):\penalty0
  280--285, 2004.
\newblock ISSN 13646613.
\newblock \doi{10.1016/j.tics.2004.04.001}.

\bibitem[Hegarty and Sims(1994)]{Hegarty1994}
Mary Hegarty and Valerie~K. Sims.
\newblock {Individual differences in mental animation during mechanical
  reasoning}.
\newblock \emph{Memory \& Cognition}, 22\penalty0 (4):\penalty0 411--430, 1994.
\newblock ISSN 0090502X.
\newblock \doi{10.3758/BF03200867}.

\bibitem[Hegarty and Steinhoff(1997)]{Hegarty1997}
Mary Hegarty and Kathryn Steinhoff.
\newblock {Individual differences in use of diagrams as external memory in
  mechanical reasoning}.
\newblock \emph{Learning and Individual Differences}, 9\penalty0 (1):\penalty0
  19--42, 1997.
\newblock ISSN 1041-6080.
\newblock \doi{0.1016/S1041-6080(97)90018-2}.

\bibitem[Hegarty et~al.(1988)Hegarty, Just, and Morrison]{Hegarty1988}
Mary Hegarty, Marcel~Adam Just, and Ian~R. Morrison.
\newblock {Mental models of mechanical systems: Individual differences in
  qualitative and quantitative reasoning}.
\newblock \emph{Cogntive Psychology}, 20\penalty0 (2):\penalty0 191--236, 1988.
\newblock \doi{10.1016/0010-0285(88)90019-9}.

\bibitem[Hu and Shu(2023)]{Hu2023}
Zhiting Hu and Tianmin Shu.
\newblock {Language models, agent models, and world models: The LAW for machine
  reasoning and planning}.
\newblock \emph{arXiv}, 2023.
\newblock \doi{10.48550/arXiv.2312.05230}.

\bibitem[Jabarian(2024)]{Jabarian2024}
Brian Jabarian.
\newblock {Black boxes: Mental models and AI models}.
\newblock In Jonathan~H Hamilton and Anindya Banerjee, editors, \emph{{Oxford
  research encyclopedia of economics and finance}}. Oxford University Press,
  2024.

\bibitem[Landis and Koch(1977)]{Landis1977}
Richard Landis and Gary Koch.
\newblock {The measurement of observer agreement for categorical data}.
\newblock \emph{Biometrics}, 33\penalty0 (1):\penalty0 159--174, 1977.
\newblock ISSN 0006341X.
\newblock \doi{10.2307/2529310}.

\bibitem[LeCun(2022)]{LeCun2022}
Yann LeCun.
\newblock {A path towards autonomous machine intelligence}.
\newblock \emph{Open Review}, \penalty0 (0.9.2):\penalty0 1--62, 2022.

\bibitem[Lewis and Mitchell(2024)]{Lewis2024}
Martha Lewis and Melanie Mitchell.
\newblock {Using counterfactual tasks to evaluate the generality of analogical
  reasoning in large language models}.
\newblock \emph{arXiv}, 2024.
\newblock \doi{10.48550/arXiv.2402.08955}.

\bibitem[Marcus(2020)]{Marcus2020}
Gary Marcus.
\newblock {The next decade in AI: Four steps towards robust artificial
  intelligence}.
\newblock \emph{arXiv}, 2020.
\newblock \doi{10.48550/arXiv.2002.06177}.

\bibitem[OpenAI(2023{\natexlab{a}})]{OpenAI2023}
OpenAI.
\newblock {GPT-4 technical report}.
\newblock \emph{arXiv}, 2023{\natexlab{a}}.
\newblock \doi{10.48550/arXiv.2303.08774}.

\bibitem[OpenAI(2023{\natexlab{b}})]{OpenAI2023a}
OpenAI.
\newblock {GPT-4V(ision) system card}.
\newblock \emph{OpenAI}, 2023{\natexlab{b}}.

\bibitem[OpenAI(2024)]{OpenAI2024}
OpenAI.
\newblock {GPT-4o System Card}.
\newblock \emph{arXiv}, 2024.
\newblock \doi{10.48550/arXiv.2303.08774}.

\bibitem[Schwartz and Black(1996)]{Schwartz1996}
Daniel~L. Schwartz and John~B. Black.
\newblock {Analog imagery in mental model reasoning: Depictive models}.
\newblock \emph{Cognitive Psychology}, 30\penalty0 (2):\penalty0 154--219,
  1996.
\newblock \doi{10.1006/cogp.1996.0006}.

\bibitem[{S{\'{e}}bastien Bubeck et al.}(2023)]{Bubeck2023}
{S{\'{e}}bastien Bubeck et al.}
\newblock {Sparks of artificial general intelligence: Early experiments with
  GPT-4}.
\newblock \emph{arXiv}, 2023.
\newblock \doi{10.48550/arXiv.2303.12712}.

\bibitem[Shepard and Metzler(1971)]{Shepard1971}
Roger~N. Shepard and Jacqueline Metzler.
\newblock {Mental rotation of three-dimensional objects}.
\newblock \emph{Science}, 171\penalty0 (3972):\penalty0 701--703, 1971.
\newblock \doi{10.1126/science.171.3972.701}.

\bibitem[Tantau(2024)]{Tantau2024}
Till Tantau.
\newblock {The TikZ and PGF Packages: Manual for version 3.1.10}, 2024.

\bibitem[Wang et~al.(2023)Wang, Gleave, Tseng, Pelrine, Belrose, Miller,
  Dennis, Duan, Levine, and Russell]{Wang2023}
Tony~T. Wang, Adam Gleave, Tom Tseng, Kellin Pelrine, Nora Belrose, Joseph
  Miller, Michael~D. Dennis, Yawen Duan, Viktor Pogrebniak~Sergey Levine, and
  Stuart Russell.
\newblock {Adversarial policies beat superhuman go AIs}.
\newblock \emph{arXiv}, page~9, 2023.
\newblock \doi{10.48550/arXiv.2211.00241}.

\bibitem[Xiang et~al.(2023)Xiang, Tao, Gu, Shu, Wang, Yang, and Hu]{Xiang2023}
Jiannan Xiang, Tianhua Tao, Yi~Gu, Tianmin Shu, Zirui Wang, Zichao Yang, and
  Zhiting Hu.
\newblock {Language models meet world models: Embodied experiences enhance
  language models}.
\newblock \emph{Advances in Neural Information Processing Systems},
  36:\penalty0 1--21, 2023.
\newblock \doi{10.48550/arXiv.2305.10626}.

\end{thebibliography}

\newpage
\section{Technical Appendices and Supplementary Material}
\label{sec:tech_extra}

\appendix
\renewcommand{\thefigure}{A\arabic{figure}}
\renewcommand{\thetable}{A\arabic{table}}
\setcounter{figure}{0}
\setcounter{table}{0}

\begin{table}[H]
	\centering
	\caption{Regression results by model, Study 1 (MA-only)}
	\label{tab:s2-ma-regression-results}
	\begin{tabular}{lcccc}
		\midrule
		& \textit{gpt-4-0314} & \textit{gpt-4-vision-preview} & \textit{claude-3-opus-20240229} & \textit{gpt-4o-2024-11-20} \\
		\midrule
		\addlinespace
		Intercept   & 0.000    & 0.000    & 0.000    & 0.000    \\
		& (0.032)  & (0.034)  & (0.034)  & (0.027)  \\
		\addlinespace
		MA          & 0.424*** & 0.200*** & 0.283*** & 0.651*** \\
		& (0.032)  & (0.034)  & (0.034)  & (0.027)  \\
		\addlinespace
		\cmidrule{2-5}
		\addlinespace
		R$^2$       & 0.180    & 0.040    & 0.080    & 0.424    \\
		Adj. R$^2$  & 0.179    & 0.039    & 0.079    & 0.423    \\
		F-statistic & 176.9*** & 33.69*** & 70.33*** & 595.0*** \\
		Residual SE & 0.906    & 0.980    & 0.960    & 0.759    \\
		DF          & 808      & 808      & 808      & 808      \\
		\midrule
		\multicolumn{5}{l}{\footnotesize Standard errors in parentheses. Signif. codes: *** p\textless0.001}
	\end{tabular}
\end{table}
\newpage

\begin{table}[H]
	\centering
	\caption{Regression results by model, Study 1 (system features included)}
	\label{tab:s2-regression-results}
	\begin{tabular}{lcccc}
		\midrule
		& \textit{gpt-4-0314} & \textit{gpt-4-vision-preview} & \textit{claude-3-opus-20240229} & \textit{gpt-4o-2024-11-20} \\
		\midrule
		\addlinespace
		Intercept          & 0.000    & 0.000    & 0.000    & 0.000    \\
		& (0.014)  & (0.016)  & (0.015)  & (0.013)  \\
		\addlinespace
		Mech. Adv.         & -0.040*  & -0.049*  & 0.035    & 0.135*** \\
		& (0.020)  & (0.023)  & (0.022)  & (0.018)  \\
		\addlinespace
		Num. code lines    & -0.006   & -0.026.  & -0.012   & 0.017    \\
		& (0.014)  & (0.016)  & (0.015)  & (0.013)  \\
		\addlinespace
		Num. pulleys       & 0.398*** & 0.203*** & 0.168*** & 0.464*** \\
		& (0.025)  & (0.027)  & (0.027)  & (0.022)  \\
		\addlinespace
		Num. ropes         & 0.082*** & 0.045*   & -0.009   & 0.035*   \\
		& (0.019)  & (0.021)  & (0.021)  & (0.017)  \\
		\addlinespace
		Pulley Rad.        & 0.027.   & 0.006    & 0.069*** & 0.032*   \\
		& (0.014)  & (0.016)  & (0.015)  & (0.013)  \\
		\addlinespace
		Rope diam.         & -0.003   & -0.015   & 0.004    & 0.006    \\
		& (0.014)  & (0.016)  & (0.015)  & (0.013)  \\
		\addlinespace
		Ceiling height     & -0.002   & 0.019    & 0.021    & 0.012    \\
		& (0.014)  & (0.016)  & (0.015)  & (0.013)  \\
		\addlinespace
		\cmidrule{2-5}
		\addlinespace
		R$^2$              & 0.185    & 0.043    & 0.041    & 0.352    \\
		Adj. R$^2$         & 0.184    & 0.041    & 0.039    & 0.351    \\
		F-statistic        & 131.3*** & 24.77*** & 24.56*** & 313.2*** \\
		Residual SE        & 0.903    & 0.979    & 0.980    & 0.806    \\
		DF                 & 4042     & 3840     & 4042     & 4041     \\
		\midrule
		\multicolumn{5}{l}{\footnotesize Standard errors in parentheses. Signif. codes: *** p\textless0.001, ** p\textless0.01, * p\textless0.05}
	\end{tabular}
\end{table}
\newpage

\begin{table}[H]
	\centering
	\caption{Classification metrics for model estimates, Studies 2 and 3} 
	\label{tab:combined_metrics} 
	\begin{tabular}{llllllll} 
		\hline
		\addlinespace
		Study & Model & Accuracy & Recall & Precision & F1 & $n_{left}$ & $n_{right}$ \\
		\addlinespace
		\hline
		\addlinespace
		\multirow[t]{4}{*}{Study 2}
		& claude-3-opus-20240229 & 0.920 & 0.866 & 0.926 & 0.895 & 40 & 110 \\
		& gpt-4-0314             & 0.801 & 0.801 & 0.801 & 0.801 & 76 & 70  \\
		& gpt-4-vision-preview   & 0.792 & 0.789 & 0.792 & 0.790 & 80 & 69  \\
		& gpt-4o-2024-11-20      & 0.713 & 0.714 & 0.714 & 0.714 & 69 & 74  \\
		\cmidrule(lr){2-8}
		& \textbf{Mean}          & \textbf{0.807} & \textbf{0.793} & \textbf{0.808} & \textbf{0.800} & \textbf{66.25} & \textbf{80.75} \\
		\addlinespace
		\hline
		\addlinespace
		\multirow[t]{2}{*}{Study 3} 
		& claude-3-opus-20240229 & 0.463 & 0.463 & 0.320 & 0.378 & 150 & 150 \\
		& gpt-4o-2024-11-20      & 0.553 & 0.553 & 0.399 & 0.464 & 150 & 150 \\
		\cmidrule(lr){2-8}
		& \textbf{Mean}          & \textbf{0.508} & \textbf{0.508} & \textbf{0.360} & \textbf{0.421} & \textbf{150} & \textbf{150} \\
		\hline
	\end{tabular}
\end{table}
\newpage

\begin{table}[H]
	\centering
	\caption{Accuracy on matched mechanical advantage predictions, Studies 2 and 3}
	\label{tab:combined-binom-results}
	\begin{tabular}{llclll}
		\midrule
		Study & Model & $n$ & Proportion correct & 95\% CI$_\text{low}$ & 95\% CI$_\text{high}$ \\
		\midrule
		\addlinespace
		\multirow[t]{4}{*}{Study 2}
		& \textit{claude-3-opus-20240229} & 150 & 0.920*** & 0.874 & 1.000 \\
		& \textit{gpt-4-0314}             & 146 & 0.801*** & 0.739 & 1.000 \\
		& \textit{gpt-4-vision-preview}   & 149 & 0.792*** & 0.730 & 1.000 \\
		& \textit{gpt-4o-2024-11-20}      & 143 & 0.713*** & 0.645 & 1.000 \\
		\cmidrule(lr){2-6}
		\addlinespace
		\multirow[t]{2}{*}{Study 3}
		& \textit{claude-3-opus-20240229} & 300 & 0.463 & 0.415 & 1.000 \\
		& \textit{gpt-4o-2024-11-20}      & 300 & 0.553*   & 0.504 & 1.000 \\
		\midrule
		\multicolumn{6}{l}{\footnotesize Exact binomial test, null $p = 0.5$. Signif. codes: *** $p < .001$, * $p < .05$}
	\end{tabular}
\end{table}
\newpage

\end{document}